# Genesis of Basic and Multi-Layer Echo State Network Recurrent Autoencoder for Efficient Data Representations

Naima Chouikhi, *Member, IEEE,* Boudour Ammar, *Member, IEEE,* Adel M. Alimi, *Senior Member, IEEE*

**Abstract**—It is a widely accepted fact that data representations intervene noticeably in machine learning tools. The more they are well defined the better the performance results are. Feature extraction-based methods such as autoencoders are conceived for finding more accurate data representations from the original ones. They efficiently perform on a specific task in terms of: 1) high accuracy, 2) large short term memory and 3) low execution time. Echo State Network (ESN) is a recent specific kind of Recurrent Neural Network which presents very rich dynamics thanks to its reservoir-based hidden layer. It is widely used in dealing with complex non-linear problems and it has outperformed classical approaches in a number of tasks including regression, classification, etc. In this paper, the noticeable dynamism and the large memory provided by ESN and the strength of Autoencoders in feature extraction are gathered within an ESN Recurrent Autoencoder (ESN-RAE). In order to bring up sturdier alternative to conventional reservoir-based networks, not only single layer basic ESN is used as an autoencoder, but also Multi-Layer ESN (ML-ESN-RAE). The new features, once extracted from ESN's hidden layer, are applied to classification tasks. The classification rates rise considerably compared to those obtained when applying the original data features. An accuracy-based comparison is performed between the proposed recurrent AEs and two variants of an ELM feed-forward AEs (Basic and ML) in both of noise free and noisy environments. The empirical study reveals the main contribution of recurrent connections in improving the classification performance results.

✦

## 1 INTRODUCTION

BUILDING feature representations [1] presents an occasion to integrate domain knowledge into data so that it can be exploited in machine learning. The setting of these representations is renowned to heavily swing the utility of the used methods. Tasks like classification and recognition [2] [3] need suitable and specific data structuring [4] [5]. For instance, images are taken in natural atmosphere. Hence, several factors such as lighting, corruptions and deformations intervene in the definition of an image [6] [7] [8].. This fact leads to images intra-class variableness [9]. Idem for signal processing as it is likely exposed to noise acquisition. While defining this variability, non-informative intra- classes should be removed and discriminating intra-classes are preserved. Data learning tools [10] appeal not only to circumvent the resorting to hand-designed data but also to encourage the employment of automatic learning. In fact, several applications are still giving manually the input data to the algorithm [11]. Feature extraction [12] is used in many domains such as natural language processing, computer vision, speech processing, etc. Up to now, there are hundreds of researchers who are focusing on hand- engineering vision, audio or text features, etc. They pay time and labor by staying years eliciting slowly specific data representations [13].

Despite the effectiveness of some feature-engineering works, one has to wonder if it would be done better. Surely, this labor-intensive work does not guarantee performance to confront new problems. The need to automatically extract better features has led to develop corresponding machine learning techniques. They tend to provide a convenient illustration of data in order to ensure promising accuracies. Machine learning-based feature extraction is a dominant subject following the achievements of deep learning [14]. Indeed, many researches attested the insightful modeling capability of Deep Neural Networks (DNNs). They have been hugely applied in various fields related to images and signal treatments [15].

Autoencoder is a kind of DNN models. It transforms an input sequence into output while minimizing as much as possible the distortion amount [16]. It is an artificial unsupervised feed-forward neural network which provides efficient data coding. It learns certain data representation so as to render a dimensionality alteration of these data (either reduction or expansion or keeping unaltered). Lately, the concept of AE was more widely indented in learning generative data models. Many variants of AEs have been developed in the literature.

For instance, denoising AE is a stochastic version of AE that performs two things. It tries to encode and preserve the inputs while undoing the impact of a corruption process (noise) applied to these input. Bengio *et al.* [1] gave a model of Denoising AE (DAE) in order to reconstruct a clean repaired input from a corrupted one. They provided therefore a new way to extract robust features. Su *et al.* [17] used a stacked denoising AE to detect frequent changes from multi-temporal images. Sun *et al.* [18] proposed a sparse AE-based approach for induction motor faults classification.

• All the authors are with REGIM-Lab: REsearch Groups in Intelligent Machines, University of Sfax, National Engineering School of Sfax (ENIS), BP 1173, Sfax, 3038, Tunisia.
Email: {naima.chouikhi, boudour.ammar, adel.alimi}@ieee.org

Vincent *et al.* [19] used sparse AE for human body detection. A plenty of other works are using AE for feature extraction in various domains. Overall, the majority of implemented AEs are based on a feed-forward fashion. Although they have get good classification results, they suffer from several deficiencies. The main drawback of AE in general is the fact that its training is based on gradient-based methods such as the back-propagation algorithm. In fact, this last is likely to converge to local minima. Also, the retro-propagation of the error signal throughout the whole network during several iterations may result to its explosion or vanishing.

In fact, Feed-forward neural networks (FNN) [20] have been well investigated and highly used since the emergence of the famous Back-Propagation (BP) algorithm [21]. Traditional BP is a first order gradient algorithm for parameter settings. It is used in Neural Networks to optimize weights values. However, BP can result on very slow convergence as well as local minimum problem [22]. Also, the error signal, when back- propagated throughout layers and neurons may either vanish or explode.

To remedy this problem, researchers have studied several methods to boost the optimality of BP-based training, such as second order optimization methods, subset selection methods or global optimization methods [23]. Although, they came up with esteemed results namely fast training speed and/or higher generalization ability, the majority of these methods are still unable to guarantee a global optimality.

Extreme Learning Machine (ELM) [24] is a popular single hidden layer feed-forward network. It is renowned for the quick learning speed and the high accuracy that it reflects. It has been implemented in multiple applications [24] [25]. In ELM, hidden nodes are randomly initiated then stay unaltered without the need of iterative tuning. The unique degree of freedom is located at the level of readout weights which relate the hidden neurons and the output ones. ELM boils down to solving fastly many problems. Inspired by the efficiency of the AE in feature representations and the effectiveness of ELM in systems modeling, researchers proposed ELM based AE (ELM-AE). In this purpose, Zhu *et al.* [26] implemented a hierarchical ELM-AE based approach to learn representations from massive unlabeled data. Wang *et al.* [27] developed a convolutional AE extreme learning machine network for 3D feature learning.

On the other hand, Recurrent Neural Networks RNNs [28] [29] are known by their capability to deal with complex non-linear problems as compared with FNNs. The recurrent feedback connections create a memory in which past captured information are stored and used to model the coming ones. This fact creates dynamism within the network.

Echo State Network [30] is a recent RNN based on a hidden random reservoir at the level of the hidden layer. It requires a reduced execution time to converge [31] [32]. Its training method is alike that of ELM. Some researchers have got to a conclusion that ESN is the recurrent form of ELM [33]. ESN has been successfully involved in several tasks such as prediction, classification, etc [34]. It is believed frequently that RNNs can perform only on temporal data as they are recurrent. However, there are many works in the literature that used recurrent architectures such as works in [35] and [36]. They studied an ESN-based approach to deal with classification problems. The efficiency of ESN dynamics provided by Reservoir Computing (RC) [37] [38] method has encouraged the idea of bestowing it as an AE. Typically, ESN is a three layered network. It has an input layer randomly connected to the next hidden layer which is the reservoir [39]. The reservoir neurons are linked to each other through a pre-determined sparse weight matrix. Only the synaptic weights from the reservoir to the output layer are trained using linear regression training. In this paper, we will extend the focus on Multi-Layer ESN [36]. Thus, a number of additional hidden weights will lead to a stronger synaptic connection between neurons of more than one reservoir. The main idea of this paper is to mix the makings of ESN (a recurrent network) and those of AE in an ESN-RAE framework to create efficient, effective and more accurate data representations. ESN is considered here as a new recurrent representational space of the original data patterns. ESN is studied as a basic and multi-layer network. It does not matter whether the data are temporal or not. What counts is the quality of the new features when extracted from the original ones via ESN-RAE.

In the literature, there is just a few researches which used the common AE to inject new data representations in ESN for classification or recognition purposes such as the work in [40]. Practically it is the first time that ESN is used as an AE. To ensure the efficiency of the proposed RAE, this latter's behavior is studied when a noise is randomly squirted into the training and / or testing databases.

The novelty of our contribution can be summarized as follows.

**1)** A novel non gradient-based recurrent ESN-RAE is proposed. This last avoids the problems of vanishing or exploding gradient. It provides a very simple training minimizing thereafter both the complexity and the execution time of the approach.

**2)** This study is extended to deep ESN with a multitude of reservoirs. Each hidden layer (reservoir) adds its own level of non-linearity that cannot be contained in a single layer. Hence, the data undergo many transformations before achieving the last one. New expressive features may appear after a series of transformations.

**3)** Adding a denoising aspect to the already proposed RAEs variants by injecting different gaussian noises into the studied datasets. The stability of the proposed RAE's behavior is studied.

The remainder of this paper is partitioned into four sections. In section 2, a thorough study about AE is performed. In section 3, Basic and Multi-layer Echo State Network AE ESN-RAE and ML-ESN-RAE approaches are described. Details about basic and ML-ESN are given too. The already proposed frameworks are applied to a number of data sets for classification purposes throughout section 4. Several simulation results are given and a comparison with ELM feed-forward AE (Basic and ML) is performed in section 4. Two cases are studied: noise free and noisy environments. Conclusions and outlook for future work are drawn by the end of the paper.

## 2 BASICS OF AUTOENCODER

Autoencoders [41] are one of the most promising tools capable of automatically furnishing efficient features from original unlabeled data.
An Autoencoder [42] is a fully connected feed-forward neural network. It is based on unsupervised learning. It is an automatic framework resulting on specific data representations. It pre-processes data to be used for classification purposes. The goal of an AE is to make input unlabeled data to be equal to the output ones. Two parts are to be distinguished in this kind of network; the encoding and the decoding phases. In the case of single-layer AE, the encoder and the decoder are gathered in a symmetrical structure (see Fig.1).

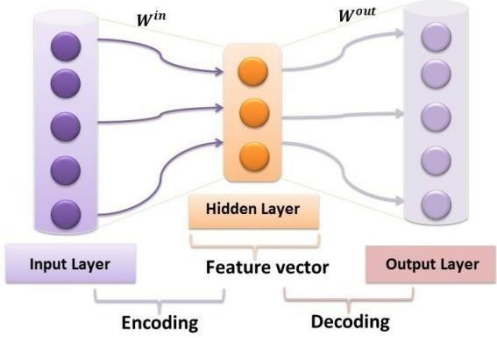

Fig. 1: Autoencoder structure.

The coding phase provides learned features extracted from the training data set. This extraction is performed following an unsupervised learning scheme. One-layer AE has one input, one output and one hidden layers. As shows Fig. 1, the input weight matrix linking the inputs to the hidden layer constitutes the encoder. The readout weights between the hidden neurons and the readout ones comprise the decoder. The dynamics of the AE are computed according to equations (1) and (2).

$$x = f_e(W^{in}u + b_e) \quad (1)$$

$$y = f_d(W^{out}x + b_d) \quad (2)$$

where $b_e$ and $b_d$ and denote the bias vectors. $f_e$ and $f_d$ represent activation functions of the encoder and the decoder, respectively. $u$, $x$ and $y$ are the input, hidden and output states. As the AEs tend to make an approximation of the identity function, the output $y$ is similar to $u$. $W^{in}$ and $W^{out}$ designate the input and output weight matrices. The training is based on Back-Propagation (BP) algorithm. Many variants of AEs are studied in the literature. They can be stacked, denoising, sparse, etc [14] [17]. It is generally recommended to make $W^{in} = (W^{out})^T$.
To remedy to the shortage of BP as well as minimize the execution time, ELM-based Autoencoders are proposed in several researches. ELM as well as ESN are recent random projections-based machine learning techniques. They have similar training algorithm based on pseudo-inverse computation of the hidden states matrix. Overall, there have not been issues focusing on using ESN as a recurrent AE. In this paper, we aim to investigate the rich dynamic of ESN's reservoir to build a competitive new variant of AEs.

## 3 BASIC AND MULTI-LAYER ECHO STATE NETWORK AUTOENCODERS

Recurrent Neural Networks (RNN) constitute an attractive kind of computational models successfully serving for nonlinear processing. To overcome RNN shortcomings especially those related to vanishing gradient, Jaeger [30] designed a specific recurrent architecture under the name of Echo State Network. This last is overall based on Reservoir Computing (RC) concept, [43]. Recently, RC has ceaselessly attracted machine learning community. Reservoir-based networks are characterized by a highly non-linear expansion provided by randomly recurrent connected neurons [44]. Basic and Multi-layer ESN variants are described in the coming subsections.

### 3.1. Basic Echo State Network Auto-encoder

#### 3.1.1. Basic Echo State Network(ESN)

The basic model of an ESN is shown in Fig. 2. Two parts are to be distinguished in such networks. The first part includes a random sparse recurrent hidden layer. It is called reservoir. It is a highly non-linear space allowing the inputs dimensions increasing. The dynamic neurons interconnected within this reservoir are activated via nonlinear function (hyperbolic tangent). The second part is a simple readout layer where the outputs are obtained by a product of the reservoir states by an output weight matrix [30] [45].

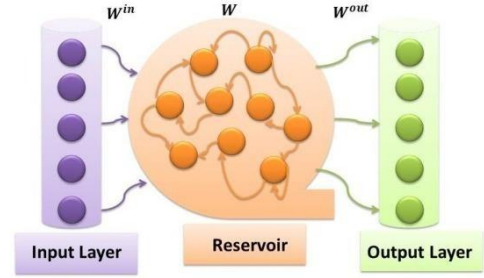

Fig. 2: Basic ESN model

Suppose $K$, $N$ and $L$ the input, hidden and output neurons, respectively. $W^{in} \in \mathsf{R}^{N*(K+1)}$ and $W \in \mathsf{R}^{N*N}$ designate the inputs plus the bias. $W^{out} \in \mathsf{R}^{L*N}$ denotes the readout weight matrix. Only these weights are to be trained.
$u(n) = [u_1(n), .., u_K(n)]^T$, $x(n) = [x_1(n), .., x_N(n)]^T$ and $y(n) = [y_1(n), .., y_L(n)]$ represent, respectively, the inputs, hidden states and outputs of the network of the $n^{th}$ pattern. $p$ is the number of patterns within the training dataset. The dynamics of ESN are represented by equations (3) and (4).

$$x(n+1) = f(W^{in} u(n+1) + W x(n)) \quad (3)$$

$$y(n+1) = f^{out}(W^{out} x(n+1)) \quad (4)$$

$f$ and $f^{out}$ are the activation functions of the reservoir and the output layer, respectively. Generally, the hidden function is non-linear while the output one is linear. Once the initial parameters are set, the training can be launched according to teacher-forced data. Thereafter, the reservoir states are aggregated in a matrix $H$. Each column in $H$ corresponds to the hidden activation states for one pattern. The output weights are computed according to equation (5) then injected to matrix *R*.

$$R = pinv(H) * y_t \qquad (5)$$

$pinv(H)$ is the pseudo-inverse of *H*. $y_t$ represents the desired outputs corresponding to the teacher-forced inputs. The network outputs *y* are calculated based on equation (4). During the test phase, inputs from testing dataset are squirted to the network. Once computed, the obtained outputs are compared to the desired ones. The result of this comparison indicates whether the reservoir is doing well his job or not.

As the Dynamic Reservoir (DR) is the key concept of ESN, it should be well designed in order to maintain a good performance of the whole network. The number of neurons as well as the connectivity rate intervene both in conceiving the structure of the DR. In conventional ESN, there are no precise formulas in choosing these parameters. There are just some recommendations to use a big number of neurons which are sparsely connected. Enlarging the reservoir as much as it is possible may give good precision results in terms of errors but it can lead to network complexity increase. Therefore, choosing optimal parameters to design a good reservoir is very crucial in order to maximize the networks accuracy while minimizing its complexity. The accuracy of the network is computed according to equation (6).

$$Minimize : ||W^{out} * x - y_t|| \qquad (6)$$

Commonly, the setting of the DR size seems to be among the most challenging tasks. In fact, the number of hidden neurons intervenes in the choice of the number of non-zero connections within the reservoir. Furthermore, it monitors the networkj s memory capacity. Many literature findings have declared that choosing a large reservoir is very likely to offer an asset in modeling complex problems [30] as well as enhancing the inputs' non-linear expansion. In fact, having randomly connected neurons improves the projection of nonlinear expansions of input signals.

Furthermore, maintaining stably rich dynamics is directly related to the sparsity degree of the hidden neurons. Sparsity rate means the number of zero weight connections whereas connectivity rate presents the number of non-zero connections. To do so, *W* is scaled by its spectral radius $\rho$ which in its turn should be lower than 1 to realize the Echo State Property ESP. According to $\rho$, whose expression figures in equation (7), there will be either densely connected neurons with smaller synaptic reservoir weights or sparsely connected units with higher weights values [46]. Stability can be ensured in both of situations meanwhile other networkjs characteristics may be altered.

$$\rho = Max\ (eigenvalues(W)) \qquad (7)$$

### 3.1.2. Basic Echo State Network Autoencoder (ESN-RAE)

In view of its achievements in dealing with several machine learning problems, ESN is used throughout this paper for creating new data representations. It plays the role of a recurrent AE.

Inspired by the concept of AE, an ESN-based Autoencoder is proposed in this section. The ESN preserves its normal structure. The outputs are set to be equal to the inputs. Basically, ESN training is performed in two phases. The first step consists of a random feature mapping of the inputs. The second is a linear parameters solving. In fact, the hidden layer, once initialized, maps the input data into another feature space. This mapping is provided by the non-linear transformation when applying the hidden activation function (see Fig. 3). ESN is distinguished by its random feature mapping as it uses in general the tanh function. Thus, it differs from other learning machine techniques which use usually kernel functions to create data features.

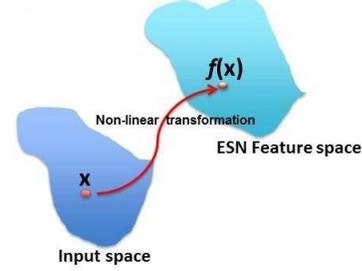

Fig. 3: ESN random feature space.

Fig. 4. gives the architecture of the basic ESN-RAE. The encoding part is composed by the inputs as well as the hidden recurrent connections. The recurrence reflected at the level of the reservoir boosts the dynamics of the network and makes it able to give more coherent data representation. The decoding phase is as usual represented by the readout activations.

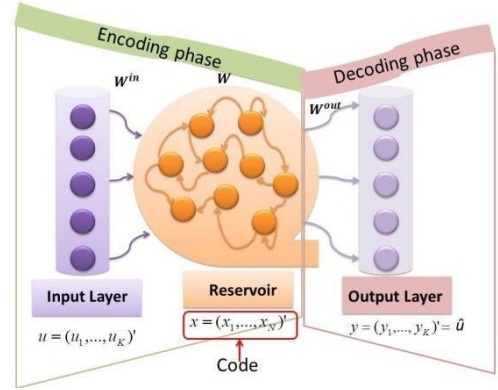

Fig. 4: Basic Echo State Network AutoEncoder (ESN-RAE).

Therefore the new equations of ESN-RAE are declared in the expressions (8) and (9).

$$x(n+1) = f_e(W^{in}u(n+1) + Wx(n) + b_e) \qquad (8)$$

$$y(n+1) = f_d(W^{out}x(n+1) + b_d) \qquad (9)$$

The output *y* is equal to *u* as we aim to approximate the inputs. The same proceeding is to be applied to ML-ESN-RAE.

### 3.2. Multi-layer Echo State Network Autoencoder

#### 3.2.1. Multi-layer Echo State Network (ML-ESN)

Very early on, researchers insisted on the idea that multi-layer networks may ensure better performance. Lately,

this idea has virtually matured thanks to the impressive successes realized across a wide range of tasks [47] [48].

Recent studies have shown that deep architectures are capable of learning complex data distributions while achieving good generalization performance and efficient representation of patterns in challenging recognition tasks [36]. Deep architecture networks have many levels of non-linearity, giving them the ability to compactly represent highly nonlinear complex mappings. However, they are difficult to train, since there are many hidden layers with many connections which cause gradient-based optimization with random initialization to get stuck in poor solutions [36].

The difference between ML-ESN and Basic ESN is the multitude of reservoirs within the network. In fact, ML-ESN is defined by a succession of interconnected reservoirs. The model of ML-ESN is visualized in Fig. 5.

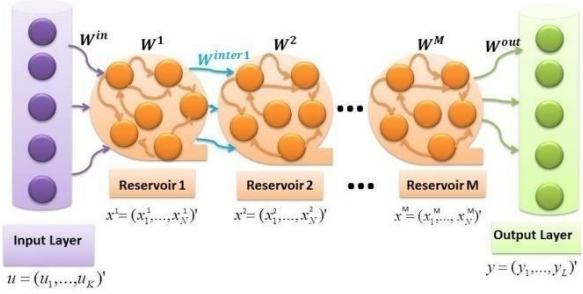

Fig. 5: ML-ESN model

The input and output layers are the same as in basic ESN. Let $M$ be the number of reservoirs in the network. Let's assume that all the reservoirs are of equal size $K$. $x^i$ designates the states vector of the $i^{th}$ reservoir, $i = 1..M$. $W^i$ denotes the inner recurrent weight matrix of reservoir $i$. $W^{inter(i)}$ defines the weights between neurons in reservoir $i$ and those of the reservoir $i + 1$. It is of size $K * K$. $W^{in}$ presents the input weight matrix relating the input neurons to those of the first reservoir. The readout weight matrix $W^{out}$ relates the last reservoir to the output layer. The dynamics of the reservoirs' states are defined in the equations (10)-(12).

$$x^1(n+1) = f(W^{in} u(n+1) + W^1 x^1(n)) \quad (10)$$

$$x^k(n+1) = f(W^{inter(k-1)} x^{k-1}(n+1) + W^k x^k(n)) \quad (11)$$

.
.
.

$$x^M(n+1) = f(W^{inter(M-1)} x^{M-1}(n+1) + W^M x^M(n)) \quad (12)$$

The output activation is computed according to equation (13).

$$y(n+1) = f^{out}(W^{out} x^M(n+1)) \quad (13)$$

The data patterns, while propagating throughout the reservoirs, undergo several non-linear transformations. Indeed, each reservoir adds a non-linear transformation of these data. These transformations may extract many new expressive features that were non-visible in their original first form.

Overall, the aim of this paper is to exploit the high non-linearity provided by the ESN js reservoir in creating another efficient representation of input patterns. Normally, the AE is used for that purpose.

### 3.2.2. Multi-layer Echo State Network Autoencoder (ML-ESN-RAE)

The architecture of ML-ESN-RAE is visualized in Fig. 6. The encoding part in ML-ESN-RAE is extended compared to that in the basic model. This time the new code is defined by the activations of the last reservoir neurons.

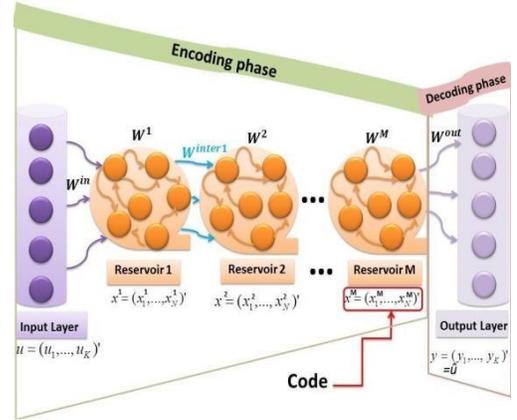

Fig. 6: Multi-layer Echo State Network Autoencoder ML-ESN-RAE.

The encoding equations of ML-ESN-RAE are expressed in equations (14)-(16).

$$x^1(n+1) = f_e (W^{in} u(n+1) + W^1 x^1(n) + b_e) \quad (14)$$

$$x^k(n+1) = f_e (W^{inter(k-1)} x^{k-1}(n+1) + W^k x^k(n) + b_e) \quad (15)$$

.
.
.

$$x^M(n+1) = f_e (W^{inter(M-1)} x^{M-1}(n+1) + W^M x^M(n) + b_e) \quad (16)$$

The decoding equation is given in (17).

$$y(n+1) = f_d(W^{out} x^M(n+1) + b_d) \quad (17)$$

where $u(n) = (u_1(n), ..., u_K(n))$ presents the input sequence of the pattern n of the database. The output $y$ is equal to $u$ as we aim to approximate the inputs.

The processing of the basic ESN is kept as it is except that we do not need ESN outputs anymore. Obtaining the reservoir states vector is the objective of this work. For each input sequence whether it was image descriptors or time series sequence, the corresponding states of the reservoir are collected in a matrix $A$. This last constitutes the new data presentation of the previous inputs. This new representation will be squirted into a classifier to fulfill the classification task.

In this work, it is hypothesized that the optimal reservoir connectivity and size depend essentially on the problem to be solved. A number of ESNs with different random weights and architectures is generated. For all the generated networks, the outputs are made to be equal to the inputs. This is to ensure the specificity of the AE. The inputs are processed to all the ESNs.

For each network, a normal learning scheme is processed till deriving $W^{out}$. The error metric is calculated for every network. The ESN (ML-ESN) architecture that provides the most reduced error value is picked out. Subsequently, the output weight matrix $W^{out}$ of the chosen network is injected to the input weight matrix to ensure the already cited condition ($W^{in} = (W^{out})^T$). It is the transposed matrix of $W^{out}$. Therefore, the input weight matrix acquires the values of the output one.

Consequently, the hidden neurons states should be recomputed because $W^{in}$ is already altered. In the case of ML-ESN, all the reservoirs activations are re-calculated too. In general, the training of overall existing feed-forward ML-AEs is done layer by layer. It means that the first hidden layer is trained first of all then its outputs are squirted as inputs to the second layer and so on till reaching the last hidden layer. This training subdivision is referred to the difficult training of these AEs. As the training of ESN is very simple and quick, we propose to train the layers at the same time.

## 4 ECHO STATE NETWORK RECURRENT AUTOENCODER APPLIED TO CLASSIFICATION

As indicates Fig. 7, the approach is mainly partitioned into two parts. The first part highlights the auto-encoding provided by the ESN-RAE (Basic and ML). This is the main phase in the approach during which a new data representation is derived from the application of input sequences to ESN-RAE. The reservoir states obtained after squirting every pattern from the dataset constitute the new representation of this pattern during the second phase.

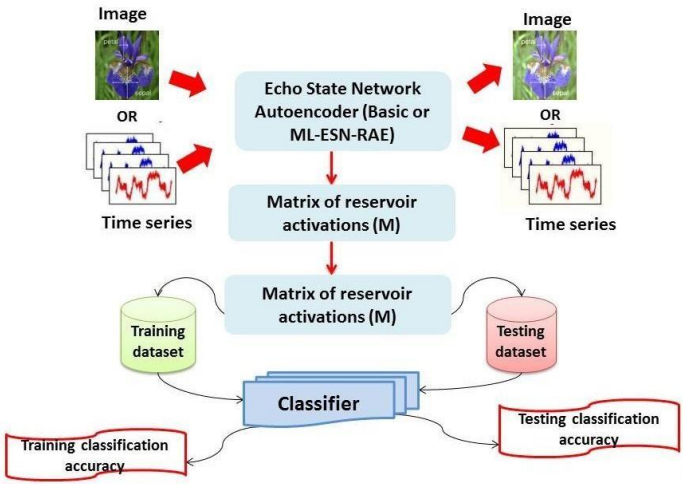

Fig. 7: Echo State Network Autoencoder framework applied to classification

During the second phase and after extracting the new data features (reservoir activations) from ESN-RAE, they are injected into a classifier.

The classification accuracy can therefore be derived. It is compared to that obtained when entering the original data directly to the classifier without passing from the ESN-RAE.

## 5 EXPERIMENTAL RESULTS

Throughout this section, a thorough empirical study of ESN-RAE (basic and ML) is carried out. The proposed framework is tested on a number of time series classification problems. Experiments are performed on a number of databases extracted from the public "UCR Time Series Data Mining Archive". These benchmarks come from various background knowledge including a broad range of application fields. The evaluation metric is represented by the classification Error Rate (ER). This last is computed according to equation (18).

$$ER = \frac{Number\ of\ Misclassified\ Patterns}{Total\ Number\ of\ Testing\ Patterns} \quad (18)$$

The performances of both basic and ML ESN-RAE frameworks are studied. For every dataset, an empirical study about the results obtained after the application of the proposed framework is given. The new data representations obtained through ESN-RAE reservoir states are passed upon a classifier. The classifier used in this work is SVM [49]. Two recurrent AEs are tested throughout this section:

- *ESN-RAE*: When original data are pre-processed by the Basic Echo State Network Autoencoder then the corresponding reservoir states are introduced to the SVM classifier.
- *ML-ESN-RAE*: When original data are pre-processed by the Multi-layer Echo State Network Autoencoder then the corresponding reservoir states are introduced to the SVM classifier.

To boost the interest on recurrent architectures, two variants of feed-forward Autoencoders are implemented to be compared to the recurrent ones. They are based on ELM as it is a feed-forward AE non trained by gradient-based methods. Also, its resemblance to ESN training and architecture lead us to implement it and show the added value of recurrent networks versus feed-forward ones. They are as follows:

- *ELM-AE*: When original data is pre-processed by the Basic Extreme Learning Machine Autoencoder then the corresponding hidden states are introduced to the SVM classifier.
- *ML-ELM-AE*: When original data is pre-processed by the Multi-layer Extreme Learning Machine Autoencoder then the corresponding hidden states of the last layer are introduced to the SVM classifier.

All of these methods are compared to each others and to other already existent literature approaches. They are executed ten times and the classification ER corresponding to each run is stored. Then, the average of the sum of ERs is taken for evaluation.

### 5.1. Datasets representation

As already said, the studied recurrent RAEs are applied to a number of time series classification datasets except the Breast cancer dataset. They are briefly described below.

### 5.1.1. ECG 200 dataset

ECG200 dataset was created by R. Olszewski [50]. It contains 200 heart beats signals are recorded. Each series or signal consists of 96 inputs. Two classes are defined here: a normal heartbeat and a Myocardial Infarction.

### 5.1.2. Breast Cancer dataset

This benchmark was generated by Dr. Wolberg [51] based on consecutive patients. Data items are extracted from digitized images. They contain illustrations of breast mass Fine Needle Aspirate (FNA). The diagnosis result will be either malignant or benign.

### 5.1.3. Coffee dataset

Two coffee bean varieties have attracted widely economic interest. They are Coffea Arabica and Coffea Canephora named also Robusta [52]. Commonly, Arabica species are more valued by the trade as they are characterized by a finer flavor.

### 5.1.4. Olive Oil dataset

Chemometrics, a task that has distinguishable applications in food safety and quality assurance, employ food spectrographs to classify food types. Each class of this data set is an extra virgin olive oil from alternative countries [52].

### 5.1.5. Earthquakes dataset

This dataset provides a prediction of whether a major earthquake is about to occur based on a set of recent readings within certain surrounding area [52].

### 5.1.6. Meat dataset

This dataset is a part of chemometrics. It is a food classification dataset. The classes are chicken, pork and turkey. Duplicate acquisitions are taken from 60 independent samples [52].

### 5.1.7. ECGFiveDays dataset

Data is from a 67 years old male [52]. The two classes are simply ECG date: 12-11-1990 ECG date: 17-11-1990.

The specifications of the datasets are gathered in Table 1 where the train and test sizes designate the size of the training and testing datasets. Length denotes the input size. The number of classes for each dataset is given too.
Seen that each dataset requires a specific ESN-RAE design,

TABLE 1: Datasets specifications

| Dataset | Train Size | Test Size | Length | Classes |
|---|---|---|---|---|
| ECG200 | 100 | 100 | 96 | 2 |
| Breast Cancer | 500 | 199 | 9 | 2 |
| Coffee | 28 | 28 | 286 | 2 |
| Earthquakes | 139 | 322 | 512 | 2 |
| Olive Oil | 30 | 30 | 570 | 4 |
| Meat | 60 | 60 | 448 | 3 |
| ECGFiveDays | 23 | 861 | 136 | 2 |

the configurations of the reservoirs used for every database figure in Table 2. $N$ and $\beta$ denote the reservoir's size and the inner connectivity, respectively. For the ML-ESN, the number of layers is chosen to be equal to 2. In fact, in [36] experiments were performed to set this number and they found that two reservoirs give better performance than just one or more than two. Also, in the same work [36], the reservoirs' are taken of equal sizes in the case of ML-ESN. In this brief, these specifications are kept.

For the size of hidden reservoirs, we were in front of two challenges. On one side, typical ESNs are based on large reservoirs. On the other side, AEs generally work on creating new compressed data representations. However, sometimes, sparse data expansion may give more accurate features. Sparsity here means that there are many features whose values are 0 or very near to zero. To deal with these issues, many tests are pre-performed before choosing the adequate hidden size. Generally, the hidden neurons in ESN are activated according to the hyperbolic tangent (tanh) function. Hence, the ESN hidden state values are in between the interval [-1,1]. Seen that the inner hidden weights are sparse, many hidden neurons activities will be very near to zero. Thus, creating large sparse representations is possible with ESN-RAE (basic or ML).

One other defiance is that in order to deal with real-world problems, a high-dimensional reservoir is usually chosen. Nevertheless, data cannot be usually obtained for a long time ago due to problems related to the shortage of data acquisition techniques [34]. Consequently, the datasets may not be so large. Here, an ill-posed problem may occur especially when the training samples number is less than the reservoir size. Significant further attention should be paid in this kind of situations.

Achieving efficient networks' specifications is a hard labor task. Table 2 represents the final settings of the proposed recurrent AEs. Taking into consideration the already cited challenges, each dataset has its specific parameters. For some of them, a compression is made (the number of hidden neurons is less than the input one). For the others, a sparse expansion is performed.

TABLE 2: Configuration of ESN-RAEs reservoir for datasets

| Parameters | $N$ | $\beta$ | $f_e$ | $f_d$ |
|---|---|---|---|---|
| ECG200 | 150 | 0.1 | tanh | linear |
| Breast cancer | 50 | 0.05 | tanh | linear |
| Coffee | 100 | 0.1 | tanh | linear |
| Olive Oil | 300 | 0.001 | tanh | linear |
| Earthquakes | 600 | 0.002 | tanh | linear |
| Meat | 250 | 0.01 | tanh | linear |
| ECGFiveDays | 100 | 0.04 | tanh | linear |

### 5.2. Accuracy results

To permit a fair comparison between recurrent and feed-forward AEs, the specification design conceived for ESN-RAE and ML-ESN-RAE is applied to ELM-AE and ML-ELM-AE. The performances of the proposed AEs are studied according to two scenarios: with and without a corruption process designated by a noise signal.

### 5.2.1. Case 1: Noise free results

As already said, creating new data features based on original ones permits sometimes to provide fitter data to the task at hand. Fig. 8 visualizes and highlights the difference between the original input data and the newly obtained features after being encoded by ML-ESN-RAE for some datasets. As a first remark, the initial input data sequences look alike. They have almost the same gait with some differences. Whereas, after being autoencoded by ML-ESN-RAE, the data get richer and more various forms regarding the original data forms. According to Fig. 8, the new obtained features (Fig.8 b),d),f)and h)) are restricted to the interval [-1,1]. The curves prove also that the obtained hidden activation states get some values nearby zero the fact that proves the sparsity aspect of the proposed ESN-RAE (Basic and ML). Tables 3-5 enumerate empirical accuracy-based results. In this case, the results are given in accordance to the classification error rate (ER). Each of the tables gives a comparison study based on the classification Error Rate (ER) between the proposed frameworks and other existent approaches that have already focused on this dataset.

The tables highlight the efficiency of using ESN-RAE and especially the impact of adding another hidden layer within the ESN for the three datasets. This is thanks to the high non-linearity provided by ESN-RAE reservoir(s). The added value of data encoding is emphasized throughout the tables as the ER when squirting the original data directly into the classifier (SVM) is higher than that obtained after their encoding. The drop surpasses 15% (for all the datasets) for ESN-RAE and 30% for ML-ESN-RAE.

TABLE 3: Classification error-based comparison with other existent approaches on the ECG200 dataset.

| Method | ER |
|---|---|
| STMF [53] | 0.300 |
| SVM [54] | 0.290 |
| LPP [55] | 0.290 |
| NCC [53] | 0.230 |
| N5S2 [56] | 0.230 |
| EDTW [55] | 0.175 |
| Jeongs Method [54] | 0.160 |
| N8S5 [56] | 0.160 |
| DSVM [55] | 0.145 |
| SVM | 0.184 |
| ELM-AE | 0.180 |
| ML-ELM-AE | 0.169 |
| *ESN-RAE* | 0.154 |
| ***ML-ESN-RAE*** | **0.113** |

TABLE 4: Classification error-based comparison with other existent approaches on the Breast cancer dataset.

| Method | ER |
|---|---|
| MLESM [36] | 0.180 |
| C-RT [51] | 0.101 |
| NB [47] | 0.070 |
| CART with feature selection [58] | 0.054 |
| C.45 [51] | 0.052 |
| SVM | 0.080 |
| ELM-AE | 0.074 |
| ML-ELM-AE | 0.063 |
| *ESN-RAE* | 0.065 |
| ***ML-ESN-RAE*** | **0.047** |

Based on what reports the tables, it comes out that ESN-RAE realizes prominent classification accuracy. ML-ESN-

TABLE 5: Classification error-based comparison with other existent approaches on the Coffee dataset.

| Method | ER |
|---|---|
| C.45 [52] | 0.178 |
| NN [52] | 0.142 |
| NB [52] | 0.142 |
| NNDTW [52] | 0.071 |
| Naive Bayes [59] | 0.071 |
| SVM | 0.106 |
| ELM-AE | 0.119 |
| ML-ELM-AE | 0.095 |
| *ESN-RAE* | 0.081 |
| ***ML-ESN-RAE*** | **0.070** |

RAE performs better than the basic one. For instance, in Table 3, ML-ESN-RAE's ER realizes a drop of more than 26% of that achieved by ESN-RAE.

According to Table 4, for Breast cancer dataset, ESN-RAE (Basic and ML) shows noticeable performances as compared to other already developed methods. The new Breast Cancer data representation based on the reservoir states has improved considerably the classification error. In this case, ML-ESN-RAE achieves an ER lessening of about 27% of ESN-RAE's ER.

Table 5 boosts the fact that Multi-layer RAE (ML-ESN-RAE) performs better than the single layer one and other proposed approaches in terms of accuracy.

So far, providing a multitude of reservoirs comes out with an enhancement in the classification accuracy according to the three precedent tables. In order to ensure further the strength of the proposed approaches, additional tests are performed on another set of benchmarks taken from the UCR archive [38]. The testing results are gathered in Table 6. Also, a comparison with existent approaches figures within the table.

From a dataset to another, the makings of the proposed

TABLE 6: Classification error-based comparison with existent methods in the literature.

| Dataset | Earthquakes | Olive Oil | Meat | ECG-Five-Days |
|---|---|---|---|---|
| ED [56] | 0.326 | 0.133 | 0.067 | 0.203 |
| DTWR [56] | 0.258 | 0.167 | 0.067 | 0.203 |
| N5S2 [56] | 0.193 | 0.567 | 0.083 | 0.220 |
| SVM | 0.309 | 0.197 | 0.103 | 0.205 |
| ELM-AE | 0.299 | 0.166 | 0.083 | 0.086 |
| ML-ELM-AE | 0.272 | 0.147 | 0.069 | 0.051 |
| *ESN-RAE* | 0.177 | 0.145 | 0.062 | 0.047 |
| ***ML-ESN-RAE*** | **0.165** | **0.124** | **0.057** | **0.028** |

ESN-RAE variants give proof of their efficiency and leverage. This is especially proved when comparing their results with those obtained when squirting the original data into the SVM classifier. This is the reason behind the insisting on the importance of all the variants of recurrent AEs conceived in this work. Overall, the tables ensure the potential of the studied ESN based AEs. Also, the use of more than one hidden reservoir influences highly the classification accuracy. In view of the results obtained so far, it appears that furnishing more than one reservoir in the hidden layer

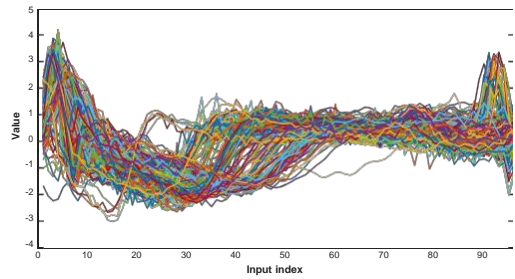
(a) Original data representation (ECG200).

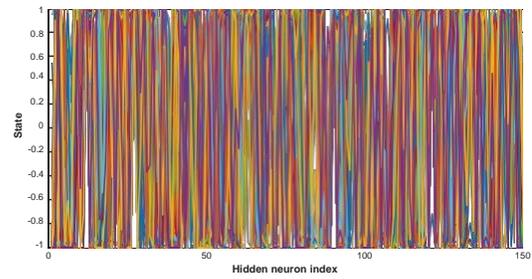
(b) New data by ML-ESN-RAE representation (ECG200).

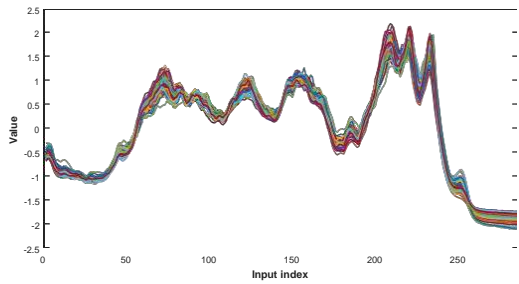
(c) Original data representation (Coffee).

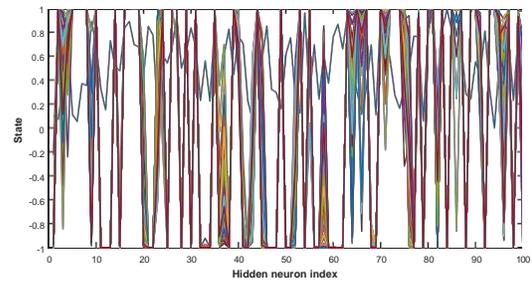
(d) New data by ML-ESN-RAE representation (Coffee).

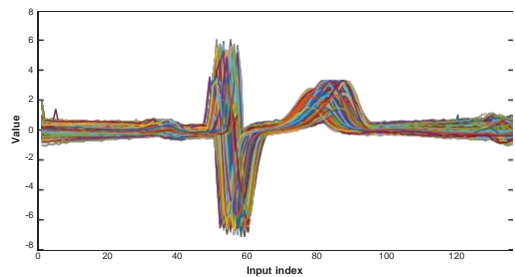
(e) Original data representation (ECGFiveDays).

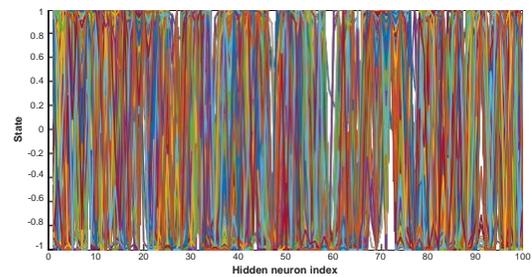
(f) New data by ML-ESN-RAE representation (ECGFiveDays).

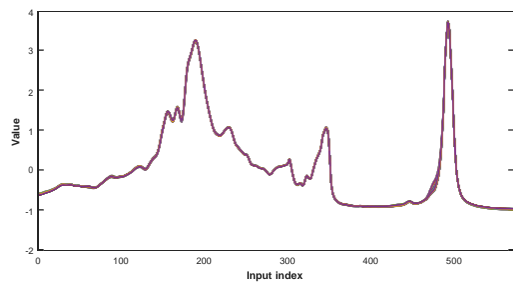
(g) Original data representation (Olive Oil).

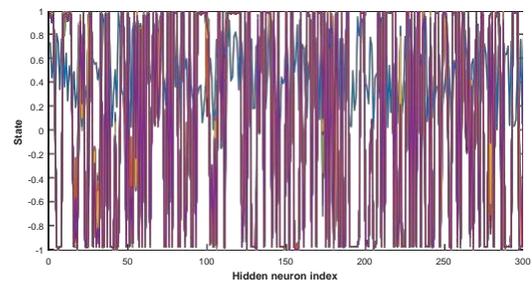
(h) New data by ML-ESN-RAE representation (Olive Oil).

Fig. 8: Original and new (by ML-ESN-RAE) datasets representations.

of ESN ensue an improvement of the AE's capabilities. For instance, the ER drops by 26%, 27%, 13%, 7%, 14%, 8% and 40% for ECG200, breast cancer, coffee, earthquakes, Olive oil, Meat and ECGFiveDays datasets, respectively, when passing from basic to ML-ESN-RAE.

Let's focus on feed-forward versus recurrent AEs performances. In fact, the classification error degrades when moving from feed-forward to recurrent fashion and from single to ML architectures. The degradation rate differs from a dataset to another. From ML-ELM-AE to ML-ESN-RAE, the ER decreases by 33%, 25%, 26%, 40%, 15%, 17% and 45% for the already cited databases. This analysis gives an idea about the efficiency of conceiving recurrent connections within the AE. Similarly, from ELM-AE to ESN-RAE the ER diminishes nearby 12% and 45%. It is to be noticed that the results achieved by ESN-RAE are better than those realized by ML-ELM-AE. Till now, the recurrent AEs give more precise and accurate results in a noise free environment. Will they preserve this advance when squirting several noise signals within the datasets? The answer to this question is to be discovered in the next paragraph.

## 5.2.2. Case 2: Gaussian noise addition

In order to study the robustness of the proposed recurrent AEs, a gaussian noise signal is squirted randomly in both of training and testing databases. Four noise levels are studied. Each noise signal is designated by a signal to noise ratio SNR. The noise level is inversely proportional to the SNR. The lower the SNR is set, the higher the noise becomes. The SNR is expressed in dB and the four levels correspond to SNR=50dB, SNR=10dB, SNR=1dB and SNR=0.5dB.

Table 7 illustrates a statistical comparison study between ESN-RAE, ML-ESN-RAE, ELM-AE and ML-ELM-AE according to the noise level. The best ER for each situation is written in bold.

It is a normal fact that the results go down at each time the amount of noise signal increases. Since the recurrent AEs are supposed to overtake the feed-forward ones in the presence of uncertainties, the accuracy evolution of the four AEs is tracked according to the amount of noise. To analyze further the effect of noise squirting, the evolution of accuracy results is visualized throughout the histograms in Fig. 9. According to Fig. 9, recurrent AEs surpass feed-forward ones in overall the cases. Even with noise supplying, the precedence of ESN-based AEs is preserved in front of feed-forward ELM-AE variants.

TABLE 7: Classification ER-based comparison between ESN-RAE, ML-ESN-RAE, ELM-AE and ML-ELM-AE with and without addition of gaussian noise levels (from top to bottom: noise free, SNR=50, SNR=10, SNR=1 and SNR=0.5)

| Method | ESN-RAE | ML-ESN-RAE | ELM-AE | ML-ELM-AE |
|---|---|---|---|---|
| ECG200 | 0.154 | **0.113** | 0.190 | 0.189 |
|  | 0.160 | **0.130** | 0.203 | 0.170 |
|  | 0.270 | **0.260** | 0.310 | 0.290 |
|  | 0.290 | **0.270** | 0.320 | 0.300 |
|  | 0.300 | **0.295** | 0.350 | 0.330 |
| Breast Cancer | 0.065 | **0.047** | 0.074 | 0.063 |
|  | 0.081 | **0.073** | 0.119 | 0.099 |
|  | 0.141 | **0.120** | 0.150 | 0.155 |
|  | **0.175** | 0.175 | 0.186 | **0.175** |
|  | 0.191 | **0.150** | 0.206 | 0.180 |
| Coffee | 0.081 | **0.070** | 0.119 | 0.095 |
|  | 0.107 | **0.071** | 0.142 | 0.108 |
|  | 0.285 | **0.193** | 0.357 | 0.321 |
|  | 0.292 | **0.221** | 0.400 | 0.357 |
|  | 0.360 | **0.321** | 0.428 | 0.364 |
| Earthquakes | 0.177 | **0.165** | 0.257 | 0.256 |
|  | 0.242 | **0.227** | 0.273 | 0.267 |
|  | 0.245 | **0.236** | 0.279 | 0.272 |
|  | 0.251 | **0.239** | 0.299 | 0.276 |
|  | 0.290 | **0.257** | 0.360 | 0.320 |
| ECGFiveDays | 0.047 | **0.028** | 0.086 | 0.051 |
|  | 0.053 | **0.034** | 0.088 | 0.061 |
|  | 0.185 | **0.179** | 0.212 | 0.199 |
|  | 0.212 | **0.198** | 0.271 | 0.204 |
|  | 0.216 | **0.209** | 0.298 | 0.229 |

In accordance with Fig. 9 and Table 7, it is quite clear that even when the noise level rises-up, the proposed approaches perform better than ELM-based AEs. For instance, in an extremely noisy environment when the SNR=0.5, the error results brought by ML-ESN-RAE detract nearby 10 to 20% (according to the studied dataset) regarding those achieved by ML-ELM-AE. Also, when SNR=50 (low noise level), the ESN-RAE achieved an ER decrease of about 20 to 40% compared to that of ELM-AE. The same thing for the other noise levels, ML-ESN-RAE is always getting the least error. Practically, ESN-RAE outperforms ML-ESN-RAE in just one case. For the other cases, ML-ESN-RAE preserves its forefront despite noise increasing. The performances of ESN-RAE and ML-ELM-AE are almost near from each others. However, ESN-RAE outstrips ELM-AE in the majority of cases.

According to Fig. 9 (f), for Meat data, the ML-ESN-RAE's ER increases by 0.043 from noise free to noisy environment with SNR=50dB. The same ER rises by 0.053 for ML-ELM-AE. As well for Olive Oil dataset where the error difference between noise free and SNR=50dB is equal to 0.143 for ML-ESN-RAE and 0.164 for ML-ELM-AE. The error rising rate when passing from a noise level to another differs from a dataset to another. Overall, in the majority of times, either ESN-RAE or ML-ESN-RAE achieves the least rate.

Let's assume the following terms: P1=ER(ML-ESN-RAE)/ER(ESN-RAE), P2=ER(ML-ESN-RAE)/ER(ML-ELM-AE) and P3=ER(ESN-RAE)/ER(ELM-AE). $P1$, $P2$ and $P3$ are computed, in Table 8, for all of noise levels in order to show thoroughly the advance of our recurrent AEs. They are multiplied by 100 in order to get the results in percentage. We can say that the ML-ESN-RAE's ER is equal to $P2$% of ML-ELM-AE's ER and $P1$% of ESN-RAE. This last, in its turn, gives an error equal to $P3$% of that reached by ELM-AE.

As shows Table 8, the difference between feed-forward and recurrent AEs (Columns of P2 and P3) seems to be remarkable as it reaches sometimes more than 40% and 30% between ML-ESN-RAE and ML-ELM-AE. Even for the single layer AEs, the recurrent one is the winner in terms of classification accuracy. Based on the obtained values of P1, using more than a single reservoir reflects a prominent improvement in terms of precision results.

TABLE 8: Percentage of ER difference between ESN-RAE, ML-ESN-RAE, ML-ELM-AE and ELM-AE with and without addition of gaussian noise levels (from top to bottom: noise free, SNR=50, SNR=10, SNR=1 and SNR=0.5)

| Dataset | P1 (%) | P2 (%) | P3 (%) |
|---|---|---|---|
| ECG200 | 73.33 | 59.78 | 81.05 |
|  | 81.25 | 76.47 | 78.81 |
|  | 96.29 | 89.65 | 80.64 |
|  | 93.10 | 90.00 | 90.62 |
|  | 98.33 | 89.39 | 85.71 |
| Breast Cancer | 72.30 | 74.60 | 87.83 |
|  | 90.12 | 73.73 | 68.06 |
|  | 85.10 | 77.41 | 94.00 |
|  | 100 | 100 | 94.08 |
|  | 78.53 | 88.33 | 92.71 |
| Coffee | 86.41 | 73.68 | 68.06 |
|  | 66.35 | 65.47 | 75.33 |
|  | 67.71 | 60.12 | 79.83 |
|  | 75.68 | 61.90 | 73.00 |
|  | 89.16 | 88.31 | 90.62 |
| Earthquakes | 93.12 | 64.45 | 68.64 |
|  | 93.80 | 85.01 | 88.64 |
|  | 96.32 | 86.76 | 90.07 |
|  | 95.21 | 86.59 | 83.94 |
|  | 88.62 | 80.31 | 90.62 |
| ECGFiveDays | 59.57 | 54.90 | 54.65 |
|  | 64.15 | 55.73 | 60.22 |
|  | 96.75 | 89.94 | 87.26 |
|  | 93.39 | 97.05 | 78.22 |
|  | 96.75 | 91.26 | 72.84 |

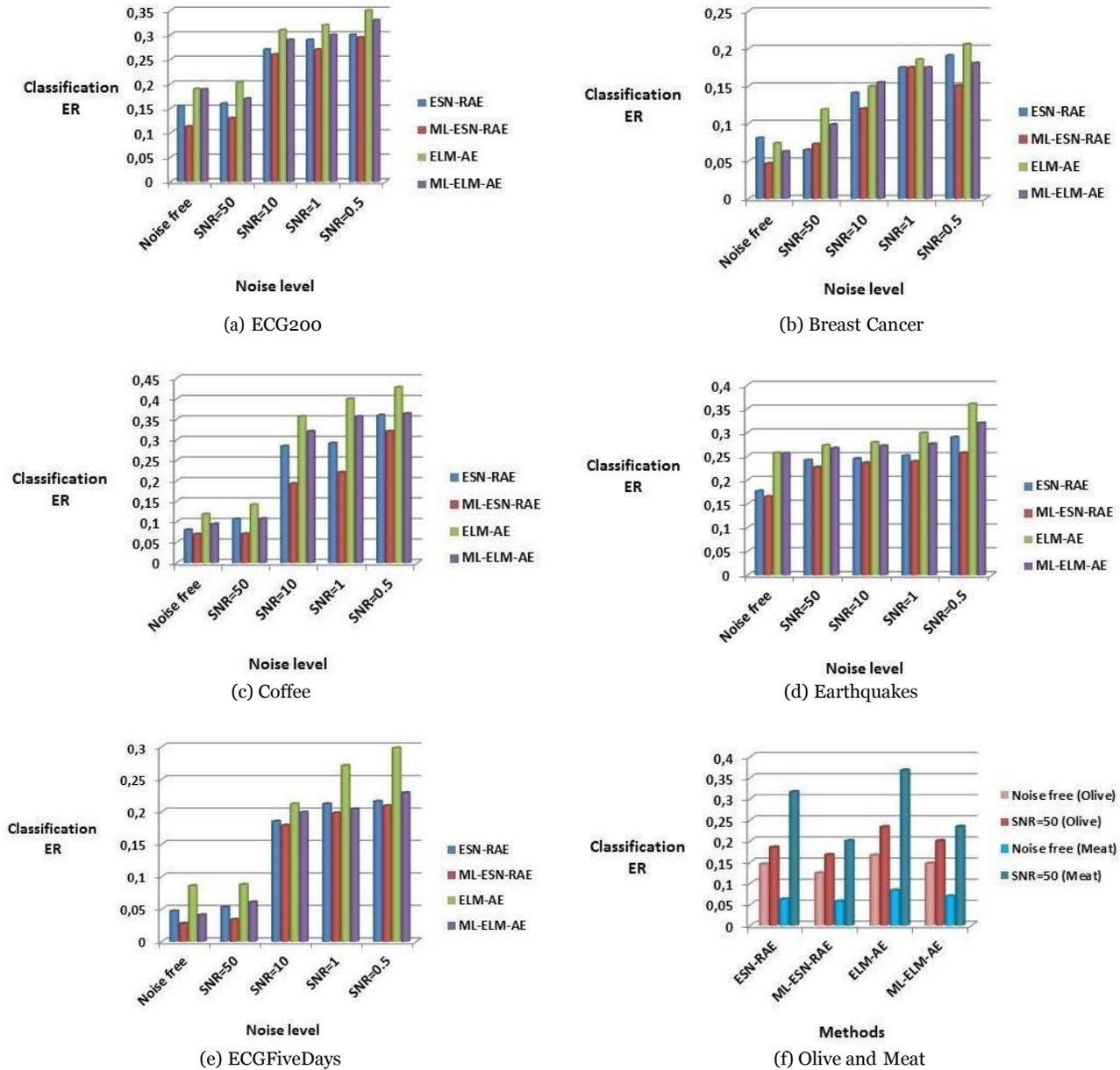

Fig. 9: Statistical analysis of the performance evolution of ESN-RAE, ML-ESN-RAE, ELM-AE, ML-ELM-AE with and without addition of gaussian noise levels.

Hence, seen what reports the precedent tables and figures, it seems that recurrent AEs have greatly better noise tolerance as they retained their robustness and advance with and without noisy data compared to other methods. Handling high and low levels of disturbance ensures the efficiency and effectiveness of the conceived recurrent ESN-AEs.

## 6 CONCLUSION

Throughout this paper, Echo State Network is exploited as an Autoencoder for data representations. The high non-linearity expansion of ESN and the strength of Autoencoder in dealing with feature extraction are merged together to bear ESN-RAE. This last is trained to provide new representation of the original data. This representation is obtained from the reservoir activation states. It encompasses other details about the data more meaningful than those of the original form. The new encoded data are squirted into a classifier. Two variants of ESN-RAE are proposed in this work which are the basic ESN-RAE and the ML-ESN-RAE. Both of ESN-RAE and ML-ESN-RAE yield considerable results in terms of classification accuracy when applied to a number of datasets. They realized an outperformance compared with existent literature approaches. In fact, the successive non-linear transformations created from a hidden layer to another may extract more significant data features. Experimental results are carried out in two kinds of environments: noise free and noisy data. A comparison study is performed between the proposed methods and other already existent approaches on a number of well-known databases. The analysis of the obtained results shows eminent potential of ESN-AE and ML-ESN-AE in terms of classification precision. Also, they reveal the importance of providing recurrent connections within the AE. This fact is proven when comparing feed-forward ELM-AE (Basic and ML) to our recurrent AEs with and without disturbance process. Indeed, the ESN-AE's variants are more powerful in terms of accuracy and stability. Other variants of ESN- RAE are under investigation to improve the results already obtained further.

## ACKNOWLEDGMENT

The research leading to these results has received funding from the Ministry of Higher Education and Scientific Research of Tunisia under the grant agreement number LR11ES48.